\documentclass[10pt,twocolumn,letterpaper]{article}

\usepackage{cvpr}
\usepackage{times}
\usepackage{epsfig}
\usepackage{graphicx}
\usepackage{amsmath}
\usepackage{amssymb}
\usepackage{diagbox}

\usepackage{multirow}
\usepackage{algorithm}
\usepackage{algpseudocode}
\usepackage{amsmath}

\usepackage[pagebackref=true,breaklinks=true,letterpaper=true,colorlinks,bookmarks=false]{hyperref}

\cvprfinalcopy 

\def\cvprPaperID{1227} 

\ifcvprfinal\fi
\begin{document}

\title{Meta-SR: A Magnification-Arbitrary Network for Super-Resolution
}

\author{Xuecai Hu\thanks{This work is conducted during Xuecai Hu's and Haoyuan Mu's internship at Megvii Inc, two authors contributed equally to this work.} $^{\, 1,2}$ , Haoyuan Mu$^{* \, 4}$, Xiangyu Zhang$^{3}$, Zilei Wang$^{1}$, Tieniu Tan$^{1, 2}$, Jian Sun$^{3}$ \\
$^{1}$ University of Science and Technology of China \\
$^{2}$ Center for Research on Intelligent Perception and Computing, NLPR, CASIA\\
$^{3}$ Megvii Inc (Face++)  $^{4}$ Tsinghua University\\
{\tt\small huxc@mail.ustc.edu.cn, muhy17@mails.tsinghua.edu.cn}\\ {\tt \small  \{zhangxiangyu, sunjian\}@megvii.com, zlwang@ustc.edu.cn, tnt@nlpr.ia.ac.cn }
}

\maketitle
\thispagestyle{empty}


\begin{abstract}

Recent research on super-resolution has achieved great success due to the development of deep convolutional neural networks (DCNNs).
However, super-resolution of arbitrary scale factor has been ignored for a  long time. Most previous researchers regard super-resolution of different  scale factors as  independent tasks. They  train a specific model for each scale factor which is inefficient in computing, and prior work only take the super-resolution of several integer scale factors into consideration.  In this work,  we propose a novel method called Meta-SR  to firstly solve  super-resolution of  arbitrary scale factor (including non-integer scale factors) with a single model. In our Meta-SR,  the Meta-Upscale Module is proposed  to replace the traditional upscale module.  For arbitrary scale factor, the Meta-Upscale Module dynamically predicts the weights of the upscale filters by taking the scale factor as input and use these weights to generate the HR image of arbitrary size. For any low-resolution image, our Meta-SR can continuously zoom in it with arbitrary scale factor by only using a single model. We evaluated the proposed method through extensive experiments on widely used benchmark datasets on single  image super-resolution.  The experimental results show the superiority of our Meta-Upscale.
\end{abstract}

\section{Introduction}
Single image super-resolution (SISR) aims to reconstruct a visually  natural high-resolution  image from its degraded low-resolution (LR) image.  And it has very wide application on  security and surveillance imaging \cite{gunturk2004super,zou2012very}, medical imaging \cite{shi2013cardiac},  as well as satellite and aerial imaging \cite{yildirim2012novel}.  In real-world scenarios, it is very common and  necessary for SISR to zoom in the LR image with  the user-customized scale factor.  As with the common image viewer, the user can arbitrarily zoom in the viewed image by rolling the mouse wheel to see the local details of the viewed image. The customized scale factor for super-resolution  also can be any positive number. And it should not be fixed to some certain integers. Thus, a  method to solve  super-resolution of arbitrary scale factor is important for putting the SISR into more practical use. If we train a specific model for each positive scale factor, it is impossible to store all these models and it is  inefficient in computing. Thus, the more important thing is
that whether we can  solve the super-resolution of arbitrary scale factor with a single  model.

However, as we all known, the most existing SISR methods only consider super-resolution of some certain integer scale factors (X2, X3, X4). And these methods treat super-resolution of different scale factors as independent tasks. Few previous work has discussed how to implement super-resolution of arbitrary scale factor. As for the state-of-the-art  SISR methods, such as  ESPCNN \cite{shi2016real}, EDSR \cite{lim2017enhanced}, RDN \cite{zhang2018residual} and RCAN \cite{zhang2018image},  these methods zoom in the feature maps at the end of networks with the sub-pixel convolution \cite{lim2017enhanced}. Unfortunately, these methods have to design a specific upscale module  for each scale factor. Each upscale module can only zoom in the image with the fixed integer scale factor. And the sup-pixel convolution only works for the integer scale factors.  These disadvantages limit the use of SISR to real-world scenarios. Although, we could implement  super-resolution of non-integer scale factors by properly upscaling the input image.  However, the repeated computation and the upscaled input make these methods very time-consuming and hard to put into pratical use.

To solve these drawbacks and put SISR into more practical use,   an efficient and novel  method for super-resolution of arbitrary scale factor with a single model is necessary.  If we want to  solve the super-resolution of arbitrary scale factor with a single model, a group of weights for upscale filters is necessary for each scale factor. Inspired by the meta-learning, we propose a network to dynamically predict the weights of  filters for each scale factor. Thus, we no longer need to store weights for each scale factor. Compared with storing the weights for each scale factor,  storing the small weight prediction network is more convenient.

We call this  method Meta-SR. There are two modules in our Meta-SR, the Feature Learning Module and the Meta-Upscale Module.  The  Meta-Upscale Module is proposed to replace the typical upscale module.
For each pixel ${(i,j)}$ on the generated HR image,  we  project it onto the LR image based on the scale factor ${r}$.  The projection coordinate is (${\lfloor \frac ir \rfloor,  \lfloor \frac jr \rfloor)}$ on the LR image.  Our Meta-Upscale Module  takes this coordinate-related and scale-related vector as input and predicts the weights for the filters.  For each pixel ${(i,j)}$ on the generated SR image,  a convolution operation is conducted between the feature at the corresponding projection coordinate on the LR image and the weights of the filters to generate the pixel value on ${(i,j)}$.
The proposed  Meta-Upscale Module could dynamically predict the variant number of  weights for the convolution filters by taking a sequence of  scale-related and coordinate-related vectors as input. Through this way , our Meta-Upscale Module can zoom in the feature maps of arbitrary scale factor with a single model.  Actually, our Meta-Upscale Module can be incorporated into most previous methods \cite{zhang2018residual, zhang2018image, lim2017enhanced} by replacing the typical upscale module.

We present extensive experiments on multiple benchmark datasets for single image super-resolution to evaluate our method. We show that: 1) For super-resolution of single integer scale factor,  our Meta-SR could achieve the comparable results with the corresponding baseline which re-trained the model for each integer scale factor. Note that our Meta-SR is trained a single model for super-resolution of arbitrary scale factor together. 2) For super-resolution of arbitrary scale factor with a single model, our Meta-SR  is  better than these methods based on properly zooming in the input images or the output images, or interpolating on the feature maps. 3) Our Meta-Upscale Module only consists of several fully connected layers and it is fast enough. The running time of our Meta-Upscale is about 1\% of the time consumed by the Feature Learning Module (RDN \cite{zhang2018residual}).

\section{Related Work}
\subsection{Single Image Super Resolution}
Early SISR methods are exemplar or dictionary based super-resolution \cite{chang2004super, timofte2013anchored,tai2017memnet}. These methods require a database of external images and generate the high-resolution images by transfering the relevant patches in the database images. The performance is limited by the size of the database or the dictionary. These traditional methods are very time-consuming and they have limited performance.

With the rapid development of the deep learning, numerous deep learning based methods have been proposed. A three-layers convolutional neural network is firstly proposed by Dong et al. \cite{dong2014learning} called SRCNN. The SRCNN upscaled the low-resolution image with bicubic interpolation before feeding into the network.  Kim et al. \cite{kim2016accurate} increased the depth of the network and used the residual learning for stable training.  Kim et al. \cite{kim2016deeply} firstly introduced the recursive learning to SISR, called DRCN. Tai et al. \cite{tai2017image} proposed DRRN by introducing the recursive blocks with shared parameters to make the training stable.
Tai et al. also introduced the memory block called Memnet \cite{tai2017memnet}.
However, the input  of  these  networks  have  the  same  size  as  the  final  high-resolution image, these methods are time-consuming.

Shi et al. \cite{shi2016real} firstly  proposed a real-time super-resolution algorithm ESPCNN by proposing the sub-pixel convolution layer. The ESPCNN \cite{shi2016real} upscaled the image at the end of the network to reduce the computation.
Ledig et al. \cite{ledig2017photo} introduced the residual block and the adversarial learning \cite{goodfellow2014generative,ganin2014unsupervised}
to make the generated images more realistic and natural. Lim et al. \cite{lim2017enhanced} used the deeper and wider residual networks called EDSR. The EDSR \cite{lim2017enhanced} removed the BN layer and used the residual scaling to speedup the training. Lim also firstly  trained single model for multiple scale factors (X2, X3, X4) called MDSR. The MDSR has different image processing blocks and upscale modules for each scale factor. Zhang et al. \cite{zhang2018residual} proposed a residual dense network (RDN) which combines the advantage of the residual block and the dense connected block. Then Zhang et al. \cite{zhang2018image} introduced the residual channel attention to the SR framework. Wang et al. \cite{Wang_2018_CVPR} proposed a novel deep spatial feature transform to recover textures conditioned on the categorical priors. Both DBPN \cite{Haris_2018_CVPR} and DSRN \cite{Han_2018_CVPR} made use of the mutual dependencies of low-  and high-resolution images. DBPN exploited iterative up-sampling and down-sampling layers to provide  an error feedback mechanism for each stage.  Jo et al. \cite{Jo_2018_CVPR} introduced the dynamic upsampling filters for video super-resolution. The dynamic upsampling filters were generated  locally and dynamically depending on the spatial-temporal neighborhood of each pixel in LR frames.  Different from this work, our  Meta-Upscale Module predicted the weights of the convolution kernel depending on the  varying scale factors for SISR. Moreover, our Meta-Upscale could generate variant number and variant weights of the convolution kernel depend on the scale factor. Instead of using the spatial-temporal feature blocks, the input of the our Meta-Upscale Module is the scale-related and coordinate-related vector. Moreover, our Meta-Upscale Module is proposed to solve the arbitrary scale.

 \begin{figure*}[]
\begin{center}
\includegraphics[width=0.95\linewidth]{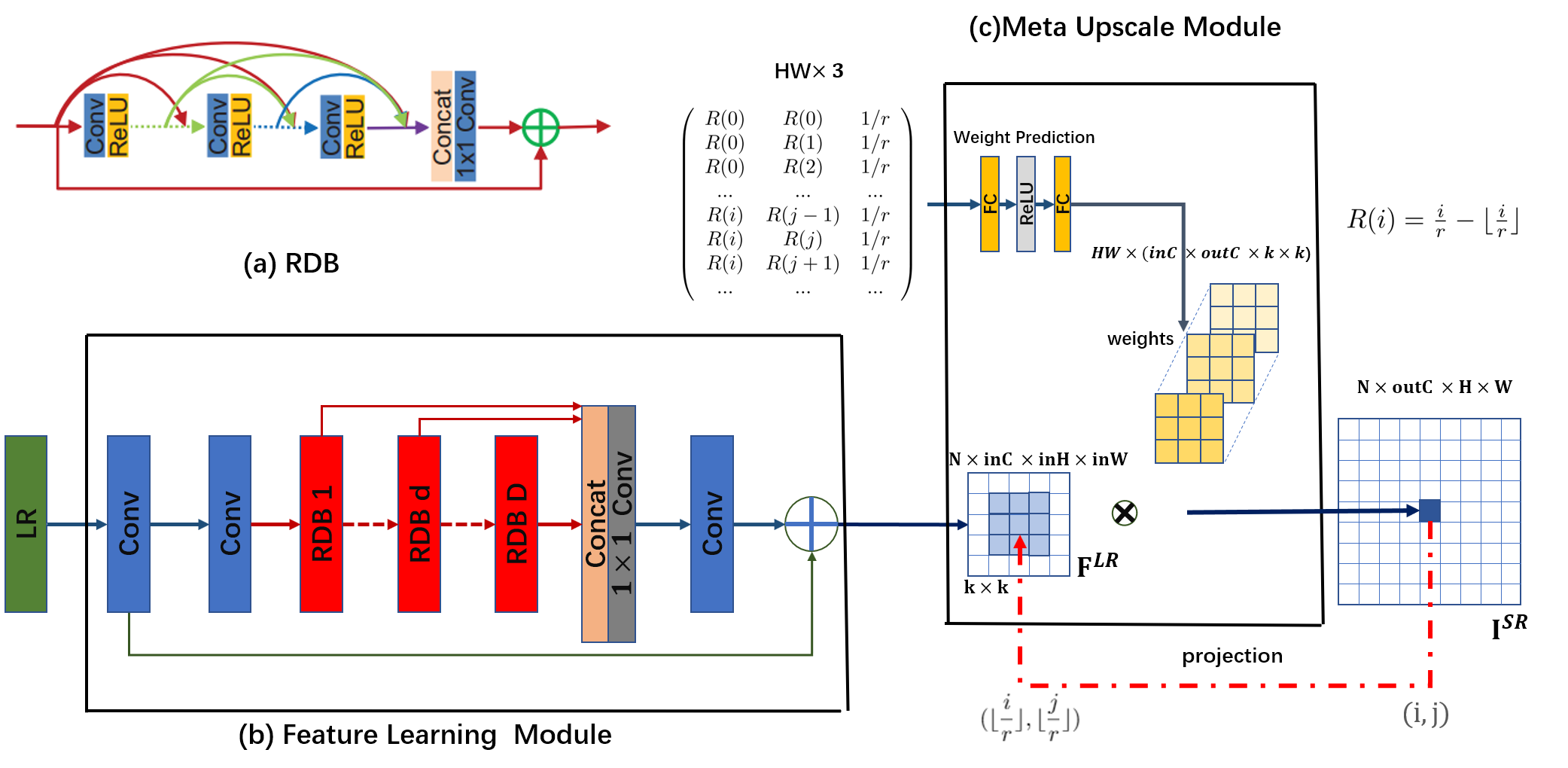}
\end{center}
   \caption{An instance  of our Meta-SR based on RDN \cite{zhang2018residual}. We also call the network Meta-RDN.  (a) The Residual Dense Block proposed by RDN \cite{zhang2018residual}.  (b) The Feature Learning Module which  generates the shared feature maps  for arbitrary scale factor. (c) For each pixel on the SR image,  we project it onto the LR image.  The proposed Meta-Upscale Module takes a sequence of  coordinate-related and scale-related vectors as input to predict the weights for convolution filters.  By doing the convolution operation, our Meta-Upscale finally generate the HR image.  }
\label{fig:ecssd3}
\end{figure*}

\subsection{Meta-Learning}
The  meta-learning, or learning to learn, is the science of observing how different machine learning  approaches perform on a wide range of learning tasks, and then learning from this experience, or meta-data. The meta-learning is mainly used in few-shot/zero-shot learning \cite{andrychowicz2016learning,ravi2016optimization} and transfer learning \cite{wang2016learning}. The detailed survey of the meta-learning can be found in \cite{lemke2015metalearning}. Here we only discuss the weight prediction related work.

The weight prediction is  one of  meta-learning strategy in the neural network \cite{lemke2015metalearning}. The weights of the neural network are predicted by another neural network rather than directly learned from the training dataset. Cai et al. \cite{Cai_2018_CVPR} predicted the parameters W of the classifier to adapt to the new categories without back propagation for few-shot learning. The parameters were predicted conditioned on the memory of  the support set. In the object detection task, Hu et al. \cite{hu2017learning} proposed to predict the mask weights from box weights. And Yang et al. \cite{yang2018metaanchor}  proposed a novel and flexible anchor mechanism for object detection. The anchor functions could be dynamically generated from the arbitrary customized prior boxes. In the video super resolution, Jo et al. \cite{Jo_2018_CVPR} proposed a dynamic upsampling filters.  The dynamic upsampling filters were generated locally and dynamically depending on the spatial-temporal neighborhood of each pixel in multiple LR frames.  Unlike this work,  we take the advantage of the meta-learning to predict weights of the filters for each scale factor. We no long need to store the weights of the filters for each scale factor. Our Meta-SR can train a single model for super-resolution of arbitrary scale. It is convenient and efficient for practical use.

The most related work is the Parameterized Image Operators \cite{fan2018decouple} which took  advantage of the weight prediction to dynamically adjust the weights of a deep network for image operators (image filtering or image restoration). Different from this work, our Meta-SR focuses on reformulation of the Upscale Module by taking both the coordinate and scale factor as inputs.


\section{Our Approach}
In this section, we describe the proposed model architectures.  As shown in Fig.\ref{fig:ecssd3}. In our Meta-SR, the Feature Learning Module extracts the feature of the low-resolution image  and the Meta-Upscale Module  upscales the feature map with arbitrary scale factor.  We  introduce our Meta-Upscale at first, then we describe the architecture details of our Meta-SR.
\subsection{Meta-Upscale Formulation}
 Given an LR image ${\textbf{I}^{LR}}$ which is downscaled from the corresponding original HR image ${\textbf{I}^{HR}}$,  the task of SISR is to generate  a HR image ${\textbf{I}^{SR}}$ whose ground-truth is ${\textbf{I}^{HR}}$. We choose the RDN \cite{zhang2018residual} as our Feature Learning Module. As shown in Fig.\ref{fig:ecssd3}(b).  Here, we focus on formulating  the Meta-Upscale Module.

 Let  ${\textbf{F}^{LR}}$ denote the feature extracted by the Feature Learning Module. Suppose  the scale factor is ${r}$. For each pixel ${(i,j)}$ on the SR image, we think that it is decided by the feature of the pixel  ${(i',j')}$ on the LR image and the weights of the corresponding filter.  From this perspective,  the upscale module can be seen as  a  mapping function to map  ${\textbf{I}^{SR}}$  and ${\textbf{F}^{LR}}$.    At first, the upscale module should   map the pixel  ${(i,j)}$ to the pixel  ${(i',j')}$.
Then,   the upscale module needs a specific filter to map the feature of the pixel ${(i',j')}$ to generate the value of this pixel ${(i,j)}$. We formulate the upscale module as:
 \begin{equation}
\label{eq:1}
\textbf{I}^{SR}(i,j) =\Phi(\textbf{F}^{LR}(i',j'), \textbf{W}(i,j))
\end{equation}
where ${\textbf{I}^{SR}(i,j)}$ denotes the pixel value at ${(i,j)}$ on SR image. ${\textbf{F}^{LR}(i',j')}$ denotes the feature  of pixel ${(i',j')}$ on the LR image. ${\textbf{W}(i,j)}$ is the weights of filter for pixel ${(i,j)}$. $ \Phi(.)$ is the feature mapping function  to calculate the pixel value.

 Since each pixel on the SR image corresponds to a filter.  For different scale factors,  both the number of the filters and the weights of the filters are different from the other scale factor.  In order to solve the super-resolution of arbitrary scale factor with a single model,  we propose the Meta-Upscale Module  to dynamically  predict the weights ${\textbf{W}(i,j)}$ based on the scale factor and coordinate information.

\begin{figure}
\begin{center}
\includegraphics[width=0.80\linewidth]{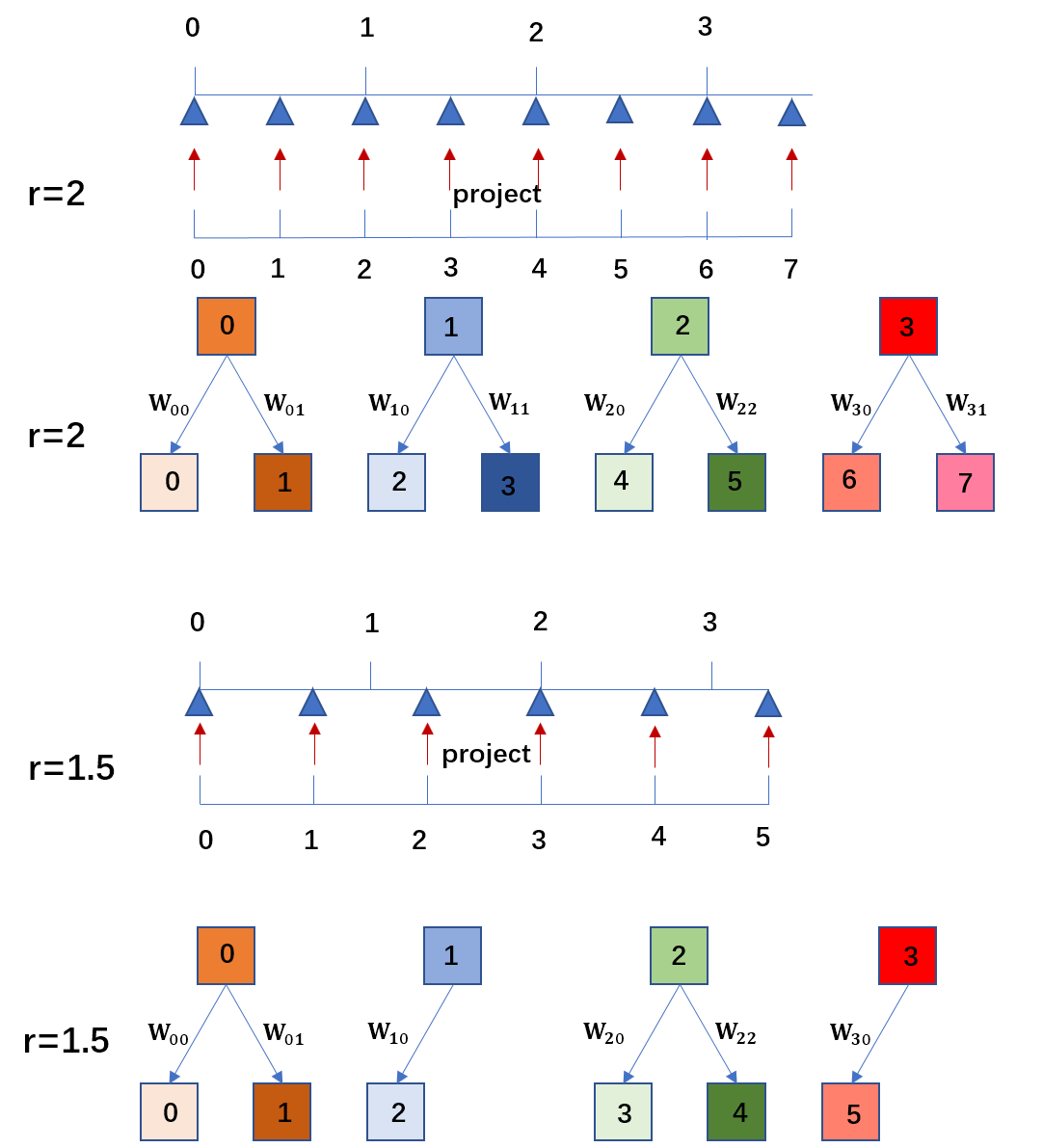}
\end{center}
   \caption{The schematic diagram for how to upscale the feature map with  the non-integer scale factor ${r=1.5}$. Here we only show the one-dimensional case for simplify.}
\label{fig:ecssd2}
\end{figure}

 For the  Meta-Upscale Module, there are three important functions.  That is, the Location Projection, the Weight Prediction and the  Feature Mapping.  As shown in the Fig.\ref{fig:ecssd2}. The Location Projection  projects  pixel onto  the LR image. And the Weight Prediction Module predicts the weights of the filter for each pixel on the SR image.
At last, the Feature Mapping function maps the feature on the LR image with the predicted weights  back to  the SR image to calculate the value of the pixel.

 \textbf{Location Projection} For each pixel ${(i,j)}$ on the SR image,
the location projection is to find the ${(i',j')}$ on the LR image. We think the value of the pixel ${(i,j)}$ is decided by  the feature of ${(i',j')}$ on the LR image. We do the following projection operator to map these two pixels:
\begin{equation}
\label{eq:1}
(i',j') = T(i,j) = \left(\left\lfloor \frac ir \right\rfloor ,  \left\lfloor \frac jr \right\rfloor\right) \\
\end{equation}
where T is the transformation function. ${\lfloor \rfloor}$ is floor function.


The Location Projection can be seen as a kind of variable fractional stride \cite{long2015fully} mechanism which could upscale the feature maps with arbitrary scale factor. As shown in the Fig \ref{fig:ecssd2}, if the scale factor ${r}$ is 2, each pixel ${(i',j')}$ determines two points. However, if the scale factor is non-integer, such as r = 1.5, some pixels determine two pixels and some pixels determine one pixel. For each pixel ${(i,j)}$  on the SR image, we could find a unique pixel ${(i',j')}$ on the LR image and we think these two pixels are most related.

 \textbf{Weight Prediction} For the typical upscale module, it predefines the number of filters for each scale factor and learns ${\textbf{W}}$ from the training dataset. Different from the typical upscale module, our Meta-Upscale Module uses a network  to  predict the weights of the filters.
We can formulate the weight prediction as:

 \begin{equation}
\label{eq:1}
\textbf{W}(i,j) = \varphi(\textbf{v}_{ij};\theta) \\
\end{equation}
 where ${\textbf{W}(i,j)}$ are the  weights of filter for pixel ${(i,j)}$ on the SR image, ${\textbf{v}_{ij}}$ is a vector related with ${i,j}$.
 ${\varphi(.)}$ is weight prediction network and takes the  ${\textbf{v}_{ij}}$  as input. ${\theta}$ is the parameters of the weight prediction network.

 As for the input of  ${\varphi(.)}$ for pixel ${(i,j)}$, the proper choice is the relative offset to the  ${(i',j')}$, the ${\textbf{v}_{ij}}$  can be formulated as:
   \begin{equation}
\label{eq:1}
\textbf{v}_{ij}= \left(\frac ir - \left\lfloor \frac ir \right\rfloor, \frac jr - \left\lfloor  \frac jr \right\rfloor\right)\\
\end{equation}

In order to train the multiple scale factor together, it is  better to add the scale factor into the ${\textbf{v}_{ij}}$ to differentiate the weights for different scale factor. For example, if we want to upscale the image with scale factor 2 and 4, and we denote them as $I^{SR}_2$ and  $I^{SR}_4$ respectively. The pixel $(i,j)$ on $I^{SR}_2$  would have the same weights of the filter  and the same projection coordinate with the pixel $(2i,2j)$ on $I^{SR}_4$. That means that  $I^{SR}_2$ is the subimage of the $I^{SR}_4$.
It would limit the performance. Thus, we redefine the ${\textbf{v}_{ij}}$  as:
\begin{equation}
\label{eq:1}
\textbf{v}_{ij}= \left(\frac ir - \left\lfloor \frac ir \right\rfloor, \frac jr - \left\lfloor  \frac jr \right\rfloor,\frac 1r\right)\\
\end{equation}

 \textbf{Feature Mapping}    We extract  the feature of the ${(i',j')}$ on the LR image from ${\textbf{F}^{LR}}$. And  we predict the weights of the filters with weight prediction network. The last thing we need to do is  mapping  feature to the value of the pixel on the SR image.  We choose the matrix product as the Feature Mapping function. We formulate the ${\Phi (.) }$as:

   \begin{equation}
\label{eq:1}
\Phi(\textbf{F}^{LR}(i',j'),\textbf{W}(i,j))=  \textbf{F}^{LR}(i',j')  \textbf{W}(i,j)\
\end{equation}

Our Meta-Upscale Module is shown in Algorithm \ref{alg::get}.

\begin{algorithm}[h]
\caption{Meta Upscale Module}
\label{alg::get}
\begin{algorithmic}[1]
\Require
scale:${r}$, the size of input image: ${(inH,inW)}$, the weight prediction function:W,  the feature: ${\textbf{F}^{LR}}$
\Ensure
  the upscale image

  \State  Calculate the output size ${outH}$ = int(${inH \times r}$), ${outW}$ = int(${inW \times r}$)
  \For{i = 0 : 1 :  ${outH}$}
      \For{j = 0 : 1 : ${outW}$}
           \State ${v_{ij}= (\frac ir - \lfloor \frac ir \rfloor, \frac jr - \lfloor  \frac jr \rfloor, \frac 1r)}$
           \State ${(i',j')=( \lfloor \frac ir \rfloor, \lfloor \frac jr \rfloor)}$
           \State  the feature on ${(i',j')}$: ${\textbf{F}^{LR}(i',j')}$
           \State  weight predicted by ${\varphi }$: ${\textbf{W}(i, j)}$
           \State ${\textbf{pv} = \textbf{F}^{LR}(i',j') \cdot \textbf{W}(i,j)}$
           \State   the  pixel value on ${(i,j)}$ is ${\textbf{pv}}$
      \EndFor
  \EndFor

  \end{algorithmic}
\end{algorithm}

\subsection{Architecture Details of Meta-SR}
There are two modules in our Meta-SR network, the Feature Learning Module, and the Meta-Upscale Module.
Most the state-of-the-art methods \cite{shi2016real,zhang2018residual,lim2017enhanced,ledig2017photo, Zhang_2018_CVPR} could be selected  as our Feature Learning Module. The proposed Meta-Upscale Module could be applied to these networks by simply replacing the traditional upscale module (sub-pixel convolution \cite{shi2016real}).
We choose the state-of-the-art SISR network, called  residual dense network ( RDN \cite{zhang2018residual} )  as   our Feature Learning Module.  Note that our Meta-SR can also work with EDSR or MDSR \cite{lim2017enhanced} or RCAN \cite{zhang2018image}.  For the RDN \cite{zhang2018residual}, there are  3 convolutional layers and 16 residual dense blocks (RDBs). Each RDB has 8 convolutional layers. The growth rate for the dense block is 64.  And the extracted feature map has   64 channels. The detailed structure  is shown in Fig.\ref{fig:ecssd3}.  More details can be found in RDN \cite{zhang2018residual}.

 For the Meta-Upscale Module,  it consists of several fully connected layers and several activation layers. Each input will output one group of  weights with the shape ${(inC, outC, k, k)}$.  Here the ${inC}$ is the number of channels of the extracted feature map, and the ${inC =64}$ in the paper.  The ${outC}$ is the number of channels of the predicted HR image. Generally, ${outC=3}$ for color images and ${outC=1}$ for grayscale image.  The ${k}$ represents the size of the convolution kernel.

 Here we want to describe the parameters of the proposed Meta-Upscale Module including the number of hidden neurons, the number of the fully connected layers, the choice of  activation function and the kernel size of the convolution layer.  Since the output size ${(k^2\times inC \times outC)}$ is very large compared with the input size (3), we set the  number of the hidden neurons as 256.  Continuing to increase the number of the hidden neurons has no improvements. And  the activation function is  ReLU. We conduct experiments and find that the best number of the fully connected layer is 2 with the balance of the speed and the performance.
As for the kernel size, ${3 \times 3}$ is the best size . Conducting ${5 \times 5}$ convolution operation on the large feature maps is more time-consuming.

\begin{table*}[h]
\caption{Results of arbitrary upscale on different methods.  The EDSR is  based on residual block. And the RDN is based on the dense connection block. The test dataset is B100 \cite{martin2001database}. The Best  results is ${\textbf{black}}$.}
\begin{center}
\begin{tabular}{|c|c|c|c|c|c|c|c|c|c|c|}
\hline

\diagbox{Methods}{Scale} &X1.1 & X1.2 &X1.3 & X1.4 &X1.5 &X1.6 &X1.7 &X1.8&X1.9&X2.0 \\
\hline
bicubic & 36.56 & 35.01 & 33.84 & 32.93 & 32.14&31.49&30.90&30.38&29.97& 29.55 \\
\hline

RDN(x1) &  42.41 &  39.76 &  38.00 &  36.68 &  35.57 &   34.64 &  33.87 &   33.19 & 32.60 & 32.08    \\

RDN(x2) & 41.84 & 39.34 & 37.87 & 36.63 & 35.56  &34.63 &33.83 &33.1 &32.52 &  32.11  \\

RDN(x4) & 39.71 & 38.48 & 37.33 & 36.29 & 35.34 &34.52&33.81 &33.14&   32.60& 32.09  \\
BiConv & 41.86 & 39.16 & 37.88 & 29.86 & 35.68  &34.77 &33.95 &33.18 &32.60 &  31.85  \\
Meta-Bi & 42.11 & 39.58 & 38.07 & 36.83 & 35.81  &34.86 &34.03 &33.24 &32.63 &  32.18  \\
Meta-RDN(\textbf{our}) & \bfseries 42.82 & \bfseries 40.40 &\bfseries 38.28 & \bfseries 36.95 &\bfseries 35.86 & \bfseries 34.90 & \bfseries 34.13 & \bfseries 33.45 & \bfseries 32.86 & \bfseries 32.35 \\
\hline
EDSR(x1) &  42.42 &  39.79 &  38.08 &  36.73 &  35.65 &  34.73  &  33.83 &  33.27 &  32.67 &   32.15  \\

EDSR(x2) & 41.79 &39.11& 37.79 & 36.51 & 35.40 & 34.49 &33.81  &33.11 &32.57 & 32.09  \\

EDSR(x4) & 39.61 & 38.41 & 37.27 & 36.24 & 35.30 &34.46  &33.75 &33.09 & 32.56  & 32.04  \\
Meta-EDSR(\textbf{our}) & \bfseries 42.72 & \bfseries 39.92 & \bfseries 38.16 & \bfseries 36.84 &  \bfseries35.78 &  \bfseries 34.83 & \bfseries 34.06 & \bfseries 33.36 & \bfseries 32.78& \bfseries 32.26\\
\hline
\hline

\diagbox{Methods}{Scale} &X2.1 & X2.2 &X2.3 & X2.4 &X2.5 &X2.6 &X2.7 &X2.8&X2.9&X3.0 \\
\hline
bicubic & 29.18 & 28.87 & 28.57& 28.31& 28.13 & 27.89 & 27.66 & 27.51 & 27.31 & 27.19 \\
\hline

RDN(x1) &  31.63 & 31.23 & 30.86 & 30.51 & 30.23 &29.95 &29.68 &29.45 &29.21& 29.03  \\
RDN(x2) & 31.61 &  31.24 & 30.82 & 30.44 & 30.23 &29.71&29.65&29.43&29.20& 29.05  \\
RDN(x4) & 31.61 & 31.23 &  30.88 &   30.52 &   30.31 &  29.99 & 29.75 &  29.53 &  29.26 &  29.14  \\
BiConv & 31.53 & 31.11 & 37.87 & 30.38 & 30.16  &29.81 &29.55 &29.28 &29.05 &  28.91  \\
Meta-Bi & 31.59 & 31.21 & 30.91 & 30.54 & 30.34  &30.01 &29.76 &29.54 &29.28 &  29.22  \\
Meta-RDN(\textbf{our}) &\bfseries 31.82 & \bfseries31.41 & \bfseries 31.06 &\bfseries  30.62 & \bfseries 30.45 &\bfseries  30.13 & \bfseries 29.82 &\bfseries  29.67 & \bfseries 29.40 & \bfseries 29.30\\
\hline
EDSR(x1) &  31.69 &  31.29 & \bfseries30.91 &  30.56 &  30.28 &   29.98  &   29.73 &   29.49 &   29.25 & 29.07  \\

EDSR(x2) &  31.57 & 31.15  & 30.81 & 30.47 & 30.22 &29.91  &29.66 &29.45 &29.19 &   29.09  \\

EDSR(x4) & 31.56 & 31.17 & 30.82 & 30.46 & 30.24 & 29.93  & 29.68 & 29.47  & 29.20 & 29.08 \\
Meta-EDSR(\textbf{our})& \bfseries 31.73 &\bfseries  31.31 &  30.87 &\bfseries  30.60 & \bfseries 30.40 & \bfseries 30.09  &\bfseries 29.83 &\bfseries 29.61 &\bfseries 29.34 & \bfseries 29.22\\
\hline
\hline

\diagbox{Methods}{Scale} &X3.1 & X3.2 &X3.3 & X3.4 &X3.5 &X3.6 &X3.7 &X3.8&X3.9&X4.0 \\
\hline
bicubic & 26.98 & 26.89 & 26.59 & 26.60 & 26.42 & 26.35 & 26.15 & 26.07 & 26.01 & 25.96 \\
\hline

RDN(x1) & 28.81 & 28.67 & 28.47 & 28.30 & 28.15 &28.00  &27.86 &27.72 &27.59 & 27.47\\
RDN(x2) & 28.71 & 28.69 & 28.51 &  28.49 & 28.18& 28.17 &\bfseries28.09&   27.84 &27.61 & 27.51  \\
RDN(x4) & \bfseries 28.89 &  28.75 &   28.57 & 28.42 &  28.19  &28.16 &27.93 &27.81  &   27.70&  27.64  \\
BiConv & 28.64 & 28.51 & 28.28 & 28.13 & 27.91  &27.84 &27.61 &27.49 &27.37 &  27.29  \\
Meta-Bi & 28.89 & 28.75 & 28.54 & 28.39 & 28.17  &28.11 &27.87 &27.75 &27.64 &  27.58  \\
Meta-RDN(\textbf{our}) &  28.87&\bfseries 28.79 &\bfseries 28.68 & \bfseries 28.54 & \bfseries 28.32 &\bfseries 28.27 &  28.04 &\bfseries 27.92 & \bfseries27.82 & \bfseries 27.75  \\
\hline
EDSR(x1) &   28.85 & 28.69 &  28.51 &   28.36 &   28.18 &28.04  &  27.91 &  27.77&   27.64 & 27.52  \\

EDSR(x2) & 28.78 & 28.64 & 28.45 & 28.34 & 28.11 &28.06  &27.83 &27.73 &27.61 & 27.49  \\

EDSR(x4) & 28.82 &   28.69 & 28.49 & 28.35 & 28.13 &   28.09  & 27.85 & 27.73 &27.63 &  27.56  \\
Meta-EDSR(\textbf{our}) &\bfseries 28.95 &\bfseries 28.82 &\bfseries 28.63 &\bfseries 28.48 &\bfseries 28.27 & \bfseries28.21  &\bfseries 27.98 & \bfseries 27.86 & \bfseries 27.75 & \bfseries 27.67\\
\hline
\end{tabular}
\end{center}
\label{table:1}
\end{table*}

\section{Experiments}
\subsection{Datasets and Metrics}
In the NTIRE 2017 Challenge on Single Image Super Resolution,   a high-quality dataset DIV2K \cite{timofte2017ntire} is newly released. There are 1000  images in DIV2K database, 800 images for training, 100 images for validation  and 100 images for test.  All of our models are trained with DIV2K training images set. For testing, we use four standard benchmark datasets: Set14 \cite{zeyde2010single}, B100 \cite{martin2001database}, Manga109 \cite{huang2015single} and DIV2K \cite{timofte2017ntire} . Note that the ground truth of the DIV2K test set is not publicly available. Therefore, we report the results on the DIV2K validation set. The super-resolution results are evaluated with PSNR and SSIM \cite{wang2004image}. Following the setting in \cite{zhang2018residual}, we only consider the PSNR and SSIM \cite{wang2004image} on the Y channel of the transformed YCbCr color space.

As for the degradation methods to generate the low-resolution images, following \cite{lim2017enhanced, zhang2018residual}, we use the bicubic interpolation by adopting the Matlab function imresize to simulate the LR images.

\subsection{Training Details}
In the single image super-resolution, the traditional loss function is L2 loss. Following the setting of \cite{lim2017enhanced}, we train our network using L1 loss instead of the L2 for better convergence.

During training the network,  we randomly extract 16 LR RGB patches with the size of 50*50 as a batch input. Following the setting in \cite{zhang2018residual}, we randomly augment the patches by flipping horizontally or vertically and rotating ${90^\circ}$. The optimizer is Adam. The learning rate is initialized to ${10^{-4}}$ for all the layers and decreases by half for every 200 epochs. All experiments  run in parallel on 4 GPUs. The training scale factors of the Meta-SR vary from 1 to 4 with stride 0.1, and the distribution of the scale factors is uniform. Each patch image in a batch has the same scale factor. Our Meta-SR is trained with Meta-Upscale Module from scratch.

\subsection{Single Model For Arbitrary Scale Factor}

Since no previous approach has focused on the  super-resolution of arbitrary scale factor with a single model,  we need to   design several baselines. We compare our approach with these baselines to prove the superiority of Meta-SR.

Suppose we want to  zoom in  the LR image with scale  ${r \in (1,4]}$. Before we feed it into the network, we could upscale it with bicubic interpolation. Thus, the first baseline simply upscales the LR image with bicubic interpolation as the final HR image, called bicubic baseline.    The second approach upscales the LR image  r times at first, and then input it into a CNN to generate the final HR image, called EDSR(x1) and RDN(x1)  respectively. These two  methods are very time-consuming and hard to put into practical use.

The third baseline  downscales the generated HR image. Suppose there is a network G to implement the ${k}$ times upscale. Thus we could input the LR image into the network G to generate the HR image. And then we downscale the HR image with scale factor ${\frac rk}$ to predict the final results.  If the ${k=2}$, we call them RDN(x2),  and EDSR(x2) respectively. For the scale ${r > k}$, we have to upscale the LR image before feeding it into the network. If the ${k=4}$, it is the fourth baseline. We call them RDN(x4) and EDSR(x4) respectively.

In order to prove the superiority of the Weight Prediction and the Location Projection, we design the fifth baseline (\textbf{BiConv}): we use the interpolation to upscale the final feature maps, the upscale module is fixed convolution layer for all scale factor. And the sixth baseline (\textbf{Meta-Bi}) is interpolating the feature maps to the needed size. We use the  Weight Prediction network to predict the weights of the convolution filter for each scale factor. We train all these models on arbitrary scale factor together.

The experimental results are shown in Table \ref{table:1}.  For the bicubic interpolation baseline, simply upscaling the LR image with bicubic interpolation could not introduce any texture or details to the HR images. It has very limited performance.  For the RDN(x1) and EDSR(x1), it has low-performance on the large scale factors. And the upscaled input makes it time-consuming.
For the RDN(x4) and EDSR(x4), the performance has huge gap between our Meta-RDN and RDN(x4) (or Meta-EDSR and EDSR(x4)) for scale factor close to 1. Moreover,  EDSR(x4) and RDN(x4) also have to upscale the LR image before feeding it into the network when the the scale ${r>k}$.

Thanks to the Weight Prediction, both Meta-Bi and our Meta-SR could learn the best weights of the filter for each scale factor while BiConv shares the same weights of the filter for all the scale factors. The experimental results show that Meta-Bi is significantly better than BiConv which proves the superiority of the Weight Prediction Module.   At the same time, our Meta-RDN is also better than the Meta-Bi. For the interpolation on the feature  maps, the larger the scale factor is, the smaller the valid Filed Of View (FOV) is. However, each scale factor has the same FOV in our Meta-SR methods.
Benefited from the proposed Meta-Upscale, our Meta-RDN  achieves the better  performance on almost all scale factors than the other baselines.

\subsection{The Inference Time}
SISR is low-level image processing task and has very high practical use. In real-world scenarios, the time requirements are very important and strict for SISR.
We measure the computing efficiency using Tesla P40 with Intel Xeon E5-2670v3@2.30GHz. We choose the B100 \cite{martin2001database} as the test dataset. Here, we do not take  the image pre-processing time into consideration.

We conduct experiments to calculate the running time of each module in our Meta-SR and the baselines. As shown in the Table.\ref{table:module}.  Compared with the Feature Learning Module, the running time of our Weight Prodiction Module can be neglected. Because there are only two fully connected layers in our Meta-Upscale Module.

Although the computing efficiency  of  our Meta-SR on single scale  has no  advantage when we compare with the baselines RDN(x1), RDN(x2) and RDN(x4) on scale ${r=2}$. If we increase the scale factor to 8 or 16, our Meta-SR is less time-consuming than these baselines.  Moreover, if we want to continuously zoom in the same image with different scale factors like the common image viewer,  our Meta-SR is the fastest. Since our Meta-SR method only need to run the Meta-Upscale Module for each scale factor. RDN(x1) and RDN(x2)  have to upscale the input image at first. And then they feed the upscaled image into whole network for each scale factor.
Thus,  we claim that our  Meta-RDN is more efficient and has better performance than these baselines.

\begin{table}[t]
\caption{Comparison running time with the baselines. FL represents the Feature Learning Module. WP is Weight Prediction Module of  our Meta-SR. Upscale is the Upscale Module. We test on the B100 and the test scale factor is 2. }
\begin{center}

\begin{tabular}{|c|c|c|c|}
\hline
Methods & FL & WP & Upscale   \\
\hline\hline
RDN(x1) & 3.66e-2s &  - & 1.7e-4s \\
RDN(x2) & 3.29e-2s &  - & 1.9e-4s \\
RDN(x4) & 3.13e-2s &  - & 3.2e-4s \\
Meta-RDN & 3.28e-2s & 1.5e-4s & 3.6e-4s \\
\hline
\end{tabular}
\end{center}
\label{table:module}
\end{table}



\subsection{Comparison With The SOTA Methods}

We apply the proposed Meta-Upscale Module to the  RDN \cite{zhang2018residual}  by replacing the typical upscale module, called Meta-RDN . We train our Meta-RDN on the DIV2K training images with random scale factor ${r \in (1,4]}$. We compare the Meta-RDN with the corresponding baseline RDN \cite{zhang2018residual}. For fair comparison, we also try to finetune our Meta-RDN for each single scale factor. However, fine-tuning on each single integer scale factor have few improvement to the final performance. And  the RDN   re-trained the model for each scale factor with different upscale module, including X2, X3, X4.
We test our  Meta-RDN on four different benchmarks with PSNR and SSIM metrics.

As shown in Table \ref{table:6},  the Meta-RDN achieve the comparable or even better results compared with the corresponding baseline RDN \cite{zhang2018residual}. Since the proposed Meta-Upscale could dynamically predict  weights of the filter for each scale and thanks to the weight prediction, we could train a single model for multiple scale factors and work well at the arbitrary scale factor.
Moreover, our Meta-SR network only need to save one model for test, but the typical model needs to save several models. Our Meta-SR network is more efficient for  SR of multiple scale factors.

\begin{table*}[]
\scriptsize
\caption{Compared with the state-of-the-art methods on X2,X3,X4. The reported results of  the state-of-the art  methods are re-trained for each scale factor. }
\begin{center}
\begin{tabular}{|c|c|c|c|c|c|c|c|c|c|c|c|c|c|}
\hline
\multirow{2}{*}{{Methods}} &
\multirow{2}{*}{Metric} &
\multicolumn{3}{c|}{Set14}&
\multicolumn{3}{c|}{B100}&
\multicolumn{3}{c|}{Manga109} &
\multicolumn{3}{c|}{DIV2K}\\
\cline{3-14}
 & &X2 &X3 &X4 &X2 &X3 &X4 &X2 &X3 &X4 &X2 &X3 &X4  \\
\hline
\multirow{2}{*}{bicubic} & PSNR  &30.24 &27.55 &26.00 &29.56 &27.21 &25.96 &30.80 &26.95 &224.89  &31.35 &28.49 &26.92  \\
& SSIM  &0.8688 &0.7742 &0.7227 &0.8431 &0.7385 &0.6675 &0.9339 &0.8556 &0.7866  &0.9076 &0.8339 &0.7774 \\
\hline
\hline

\multirow{2}{*}{RDN} & PSNR  &34.01 &\bfseries30.57 &28.81 &32.34 &29.26 &27.72 &\bfseries39.18 &34.13 &31.00   &35.17 &31.39 &29.34 \\
& SSIM  &0.9212 &\bfseries0.8468 &0.7871 &0.9017 &0.8093 &0.7419 &0.9780 &\bfseries0.9484 &0.9151   &0.9483 &0.8931 &0.8446 \\
\hline
\multirow{2}{*}{Meta-RDN} & PSNR  &\bfseries34.04 &30.55 &\bfseries28.84 &\bfseries32.35 &\bfseries29.30 &\bfseries27.75 &39.18 &\bfseries34.14 &\bfseries31.03  &\bfseries35.18 &\bfseries31.42 &\bfseries29.36 \\
& SSIM  &\bfseries0.9213 &0.8466 &\bfseries0.7872 &\bfseries0.9019 &\bfseries0.8096 &\bfseries0.7423 &\bfseries0.9782 &0.9483 &\bfseries0.9154  &\bfseries0.9484 &\bfseries0.8935 &\bfseries0.8448 \\
\hline
\end{tabular}
\end{center}
\label{table:6}
\end{table*}

 \begin{figure*}[t]
\begin{center}
\includegraphics[width=0.88\linewidth]{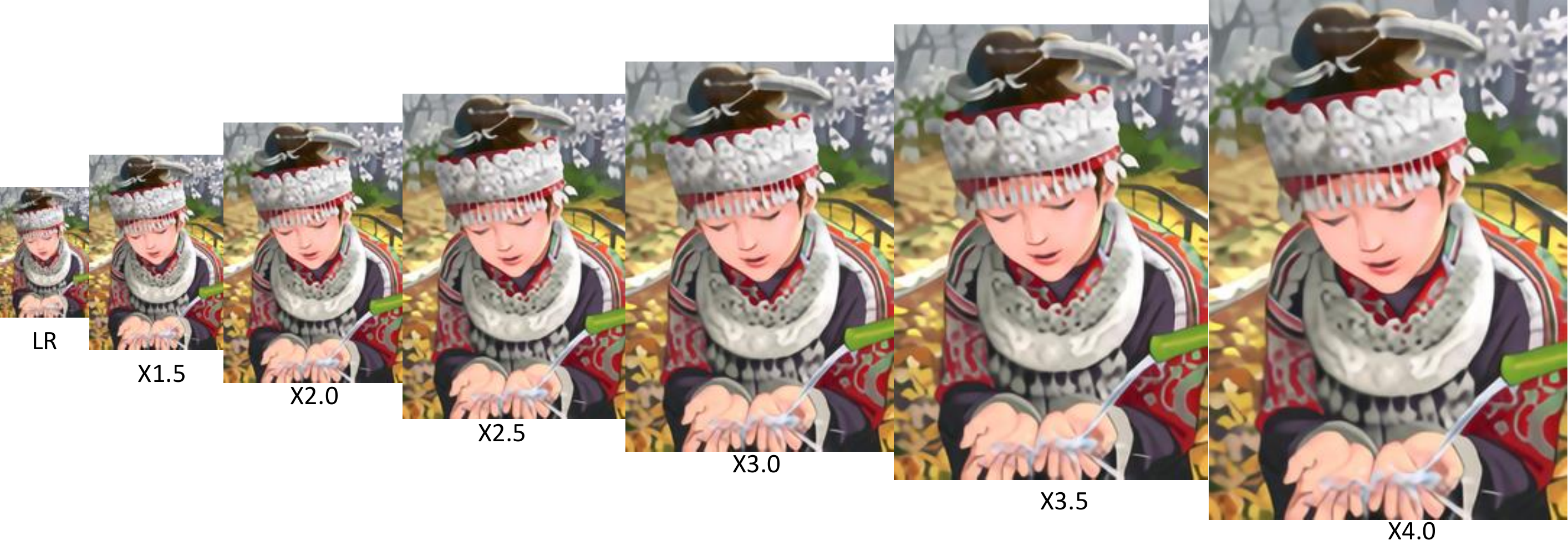}
\end{center}
   \caption{The visual comparison of an single image upsampling with different scale factors by our Meta-RDN.}
\label{fig:7}
\end{figure*}

\subsection{Visual Results}
\begin{figure}[t]
\begin{center}
\includegraphics[width=0.95\linewidth]{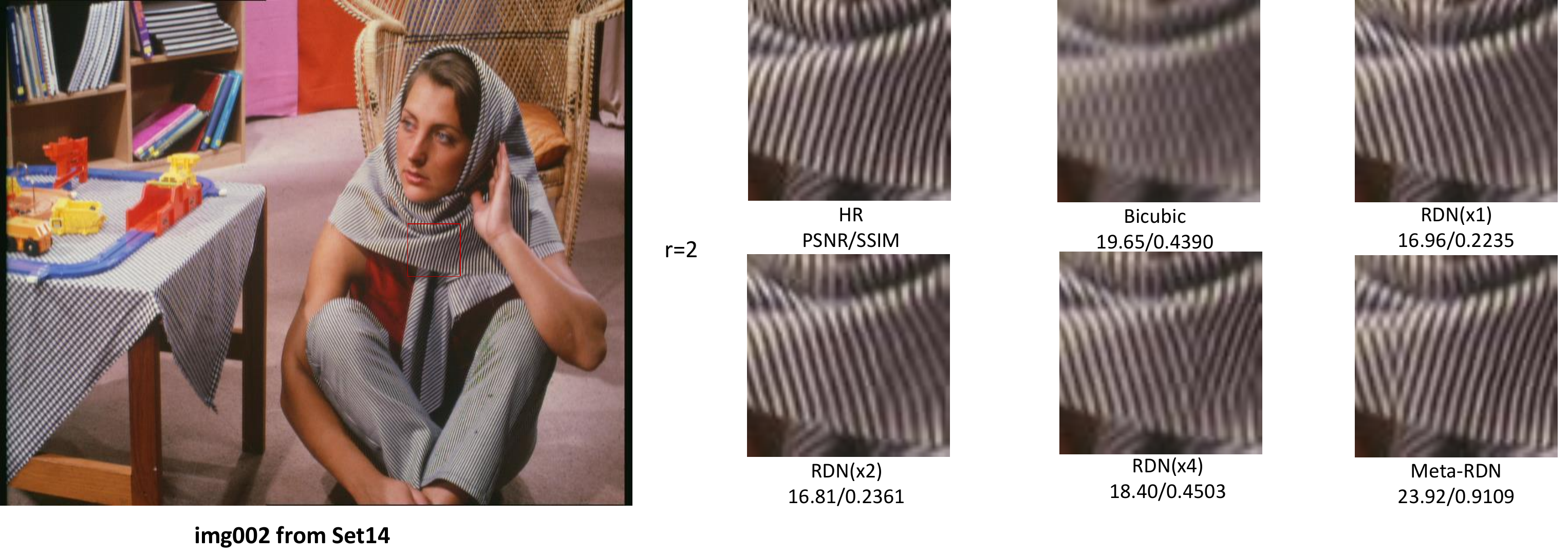}
\includegraphics[width=0.95\linewidth]{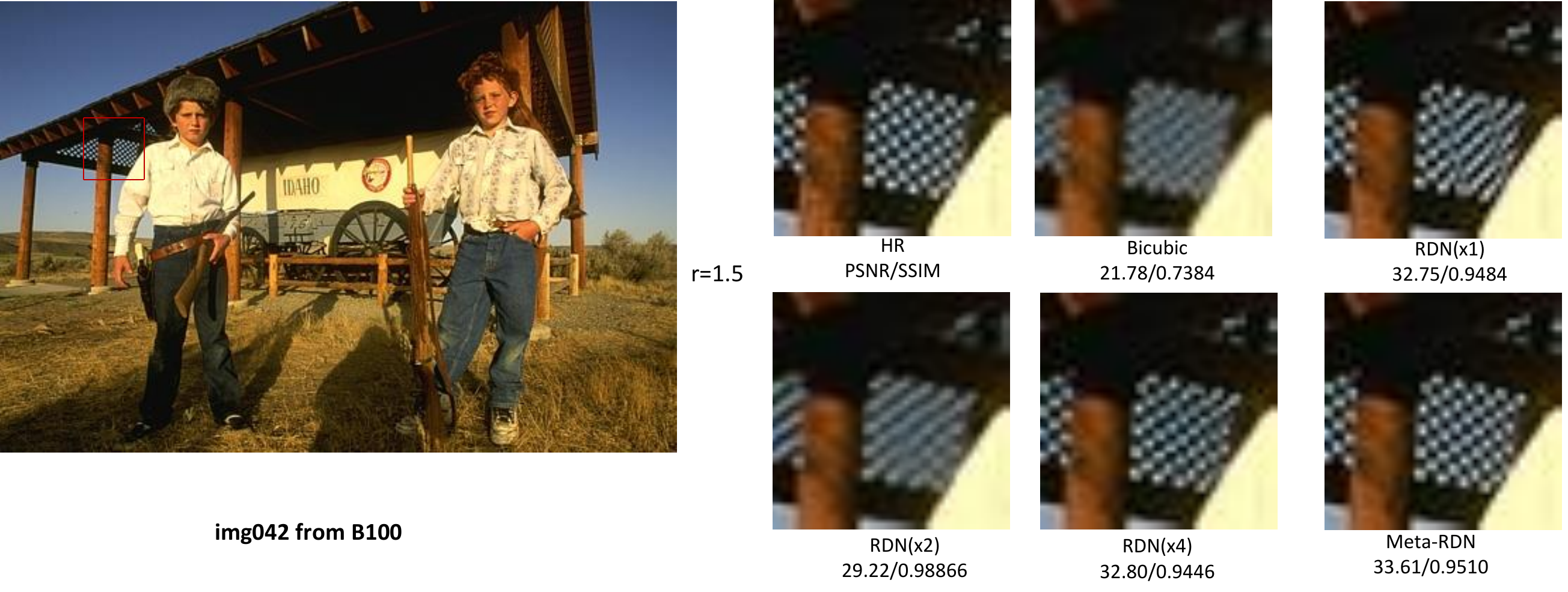}
\end{center}
   \caption{The visual comparison with  four baselines. Our Meta-RDN has better performance.}
\label{fig:8}
\end{figure}

In this section, we show the visual results in Fig.\ref{fig:7} and Fig.\ref{fig:8} .
As shown in Fig.\ref{fig:8}, we compare with the RDN(x1), RDN(x2) and RDN(x4) for super-resolution of arbitrary scale factor.  Our Meta-SR has better performance for the structure part. Since the RDN(x1), RDN(x2) and RDN(x4) share the same the weights of filters for all the scale factors, the texture of the SR image generated by these baseline methods is worse than our Meta-RDN. Thanks to the weight prediction,  our Meta-SR can predict a group of independent weights for each scale factor.

\section{Conclusion}
We propose a novel  upscale module named Meta-Upscale to solve the super-resolution of  arbitrary scale factor with a single model. The proposed Meta-Upscale Module  could dynamically predict the weights of the filters. For each scale factor, the proposed Meta-Upscale Module  generates a group of weights for the upscale module.  By doing convolution operation between the feature maps and the filters, we  generate the HR image of arbitrary size. Thanks to the weight prediction, we can train a single model for super-resolution of  arbitrary scale factor. Especially, our Meta-SR can continuously zoom in the same image with multiple scale factors.

\section*{Acknowledgement}
This research is  supported by National Key R\&D Program of China (2017YFA0700800, 2016YFB1001002, 2016YFB1001000), National Natural Science Foundation of China (61525306,
61633021, 61721004, 61420106015), Capital Science and Technology
Leading Talent Training Project (Z181100006318030), and
Beijing Science and Technology Project (Z181100008918010).

\clearpage

{\small
\bibliographystyle{ieee}
\bibliography{egbib}

\begin{thebibliography}{10}\itemsep=-1pt

\bibitem{andrychowicz2016learning}
M.~Andrychowicz, M.~Denil, S.~Gomez, M.~W. Hoffman, D.~Pfau, T.~Schaul,
  B.~Shillingford, and N.~De~Freitas.
\newblock Learning to learn by gradient descent by gradient descent.
\newblock In {\em Advances in Neural Information Processing Systems}, 2016.

\bibitem{Cai_2018_CVPR}
Q.~Cai, Y.~Pan, T.~Yao, C.~Yan, and T.~Mei.
\newblock Memory matching networks for one-shot image recognition.
\newblock In {\em The IEEE Conference on Computer Vision and Pattern
  Recognition (CVPR)}, June 2018.

\bibitem{chang2004super}
H.~Chang, D.-Y. Yeung, and Y.~Xiong.
\newblock Super-resolution through neighbor embedding.
\newblock In {\em Computer Vision and Pattern Recognition, 2004. CVPR 2004.
  Proceedings of the 2004 IEEE Computer Society Conference on}, volume~1, pages
  I--I. IEEE, 2004.

\bibitem{dong2014learning}
C.~Dong, C.~C. Loy, K.~He, and X.~Tang.
\newblock Learning a deep convolutional network for image super-resolution.
\newblock In {\em European conference on computer vision}. Springer, 2014.

\bibitem{fan2018decouple}
Q.~Fan, D.~Chen, L.~Yuan, G.~Hua, N.~Yu, and B.~Chen.
\newblock Decouple learning for parameterized image operators.
\newblock In {\em Proceedings of the European Conference on Computer Vision
  (ECCV)}, pages 442--458, 2018.

\bibitem{ganin2014unsupervised}
Y.~Ganin and V.~Lempitsky.
\newblock Unsupervised domain adaptation by backpropagation.
\newblock {\em arXiv preprint arXiv:1409.7495}, 2014.

\bibitem{goodfellow2014generative}
I.~Goodfellow, J.~Pouget-Abadie, M.~Mirza, B.~Xu, D.~Warde-Farley, S.~Ozair,
  A.~Courville, and Y.~Bengio.
\newblock Generative adversarial nets.
\newblock In {\em Advances in neural information processing systems}, pages
  2672--2680, 2014.

\bibitem{gunturk2004super}
B.~K. Gunturk, Y.~Altunbasak, and R.~M. Mersereau.
\newblock Super-resolution reconstruction of compressed video using
  transform-domain statistics.
\newblock {\em IEEE Transactions on Image Processing}.

\bibitem{Han_2018_CVPR}
W.~Han, S.~Chang, D.~Liu, M.~Yu, M.~Witbrock, and T.~S. Huang.
\newblock Image super-resolution via dual-state recurrent networks.
\newblock In {\em The IEEE Conference on Computer Vision and Pattern
  Recognition (CVPR)}, June 2018.

\bibitem{Haris_2018_CVPR}
M.~Haris, G.~Shakhnarovich, and N.~Ukita.
\newblock Deep back-projection networks for super-resolution.
\newblock In {\em The IEEE Conference on Computer Vision and Pattern
  Recognition (CVPR)}, June 2018.

\bibitem{hu2017learning}
R.~Hu, P.~Doll{\'a}r, K.~He, T.~Darrell, and R.~Girshick.
\newblock Learning to segment every thing.
\newblock {\em Cornell University arXiv Institution: Ithaca, NY, USA}, 2017.

\bibitem{huang2015single}
J.-B. Huang, A.~Singh, and N.~Ahuja.
\newblock Single image super-resolution from transformed self-exemplars.
\newblock In {\em Proceedings of the IEEE Conference on Computer Vision and
  Pattern Recognition}, 2015.

\bibitem{Jo_2018_CVPR}
Y.~Jo, S.~Wug~Oh, J.~Kang, and S.~Joo~Kim.
\newblock Deep video super-resolution network using dynamic upsampling filters
  without explicit motion compensation.
\newblock In {\em The IEEE Conference on Computer Vision and Pattern
  Recognition (CVPR)}, June 2018.

\bibitem{kim2016accurate}
J.~Kim, J.~Kwon~Lee, and K.~Mu~Lee.
\newblock Accurate image super-resolution using very deep convolutional
  networks.
\newblock In {\em Proceedings of the IEEE conference on computer vision and
  pattern recognition}, 2016.

\bibitem{kim2016deeply}
J.~Kim, J.~Kwon~Lee, and K.~Mu~Lee.
\newblock Deeply-recursive convolutional network for image super-resolution.
\newblock In {\em Proceedings of the IEEE conference on computer vision and
  pattern recognition}, 2016.

\bibitem{ledig2017photo}
C.~Ledig, L.~Theis, F.~Husz{\'a}r, J.~Caballero, A.~Cunningham, A.~Acosta,
  A.~P. Aitken, A.~Tejani, J.~Totz, Z.~Wang, et~al.
\newblock Photo-realistic single image super-resolution using a generative
  adversarial network.
\newblock In {\em CVPR}, 2017.

\bibitem{lemke2015metalearning}
C.~Lemke, M.~Budka, and B.~Gabrys.
\newblock Metalearning: a survey of trends and technologies.
\newblock {\em Artificial intelligence review}, 2015.

\bibitem{lim2017enhanced}
B.~Lim, S.~Son, H.~Kim, S.~Nah, and K.~M. Lee.
\newblock Enhanced deep residual networks for single image super-resolution.
\newblock In {\em The IEEE conference on computer vision and pattern
  recognition (CVPR) workshops}, 2017.

\bibitem{long2015fully}
J.~Long, E.~Shelhamer, and T.~Darrell.
\newblock Fully convolutional networks for semantic segmentation.
\newblock In {\em CVPR}, 2015.

\bibitem{martin2001database}
D.~Martin, C.~Fowlkes, D.~Tal, and J.~Malik.
\newblock A database of human segmented natural images and its application to
  evaluating segmentation algorithms and measuring ecological statistics.
\newblock In {\em Computer Vision, 2001. ICCV 2001. Proceedings. Eighth IEEE
  International Conference on}, volume~2, pages 416--423. IEEE, 2001.

\bibitem{ravi2016optimization}
S.~Ravi and H.~Larochelle.
\newblock Optimization as a model for few-shot learning.
\newblock 2016.

\bibitem{shi2016real}
W.~Shi, J.~Caballero, F.~Husz{\'a}r, J.~Totz, A.~P. Aitken, R.~Bishop,
  D.~Rueckert, and Z.~Wang.
\newblock Real-time single image and video super-resolution using an efficient
  sub-pixel convolutional neural network.
\newblock In {\em Proceedings of the IEEE Conference on Computer Vision and
  Pattern Recognition}, 2016.

\bibitem{shi2013cardiac}
W.~Shi, J.~Caballero, C.~Ledig, X.~Zhuang, W.~Bai, K.~Bhatia, A.~M. S.~M.
  de~Marvao, T.~Dawes, D.~O’Regan, and D.~Rueckert.
\newblock Cardiac image super-resolution with global correspondence using
  multi-atlas patchmatch.
\newblock In {\em International Conference on Medical Image Computing and
  Computer-Assisted Intervention}. Springer, 2013.

\bibitem{tai2017image}
Y.~Tai, J.~Yang, and X.~Liu.
\newblock Image super-resolution via deep recursive residual network.
\newblock In {\em Proceedings of the IEEE Conference on Computer Vision and
  Pattern Recognition}, volume~1, page~5, 2017.

\bibitem{tai2017memnet}
Y.~Tai, J.~Yang, X.~Liu, and C.~Xu.
\newblock Memnet: A persistent memory network for image restoration.
\newblock In {\em Proceedings of the IEEE Conference on Computer Vision and
  Pattern Recognition}, pages 4539--4547, 2017.

\bibitem{timofte2017ntire}
R.~Timofte, E.~Agustsson, L.~Van~Gool, M.-H. Yang, L.~Zhang, B.~Lim, S.~Son,
  H.~Kim, S.~Nah, K.~M. Lee, et~al.
\newblock Ntire 2017 challenge on single image super-resolution: Methods and
  results.
\newblock In {\em Computer Vision and Pattern Recognition Workshops (CVPRW),
  2017 IEEE Conference on}, pages 1110--1121. IEEE, 2017.

\bibitem{timofte2013anchored}
R.~Timofte, V.~De~Smet, and L.~Van~Gool.
\newblock Anchored neighborhood regression for fast example-based
  super-resolution.
\newblock In {\em Proceedings of the IEEE international conference on computer
  vision}, pages 1920--1927, 2013.

\bibitem{Wang_2018_CVPR}
X.~Wang, K.~Yu, C.~Dong, and C.~Change~Loy.
\newblock Recovering realistic texture in image super-resolution by deep
  spatial feature transform.
\newblock In {\em The IEEE Conference on Computer Vision and Pattern
  Recognition (CVPR)}, June 2018.

\bibitem{wang2016learning}
Y.-X. Wang and M.~Hebert.
\newblock Learning to learn: Model regression networks for easy small sample
  learning.
\newblock In {\em European Conference on Computer Vision}, pages 616--634.
  Springer, 2016.

\bibitem{wang2004image}
Z.~Wang, A.~C. Bovik, H.~R. Sheikh, and E.~P. Simoncelli.
\newblock Image quality assessment: from error visibility to structural
  similarity.
\newblock {\em IEEE transactions on image processing}, 13(4):600--612, 2004.

\bibitem{yang2018metaanchor}
T.~Yang, X.~Zhang, W.~Zhang, and J.~Sun.
\newblock Metaanchor: Learning to detect objects with customized anchors.
\newblock {\em NIPS}, 2018.

\bibitem{yildirim2012novel}
D.~Y{\i}ld{\i}r{\i}m and O.~G{\"u}ng{\"o}r.
\newblock A novel image fusion method using ikonos satellite images.
\newblock {\em Journal of Geodesy and Geoinformation}, 2012.

\bibitem{zeyde2010single}
R.~Zeyde, M.~Elad, and M.~Protter.
\newblock On single image scale-up using sparse-representations.
\newblock In {\em International conference on curves and surfaces}, pages
  711--730. Springer, 2010.

\bibitem{Zhang_2018_CVPR}
K.~Zhang, W.~Zuo, and L.~Zhang.
\newblock Learning a single convolutional super-resolution network for multiple
  degradations.
\newblock In {\em The IEEE Conference on Computer Vision and Pattern
  Recognition (CVPR)}, June 2018.

\bibitem{zhang2018image}
Y.~Zhang, K.~Li, K.~Li, L.~Wang, B.~Zhong, and Y.~Fu.
\newblock Image super-resolution using very deep residual channel attention
  networks.
\newblock {\em arXiv preprint arXiv:1807.02758}, 2018.

\bibitem{zhang2018residual}
Y.~Zhang, Y.~Tian, Y.~Kong, B.~Zhong, and Y.~Fu.
\newblock Residual dense network for image super-resolution.
\newblock In {\em The IEEE Conference on Computer Vision and Pattern
  Recognition (CVPR)}, 2018.

\bibitem{zou2012very}
W.~W. Zou and P.~C. Yuen.
\newblock Very low resolution face recognition problem.
\newblock {\em IEEE Transactions on Image Processing}, 2012.

\end{thebibliography}
}

\end{document}